\documentclass[3p,twocolumn]{elsarticle}

\journal{Trends in Chemistry}
\usepackage{amssymb}
\usepackage{booktabs}
\usepackage{enumitem}
\usepackage{hyperref}
\usepackage{parskip}

\begin{document}

\begin{frontmatter}

\title{Pursuing a Prospective Perspective}

\author{Steven Kearnes\corref{mycorrespondingauthor}}
\address{Google Research, Applied Science Team \\ Mountain View, California 94043, USA}
\cortext[mycorrespondingauthor]{Correspondence: \href{mailto:kearnes@google.com}{\texttt{kearnes@google.com}} (Steven Kearnes)}

\begin{abstract}
Retrospective testing of predictive models does not consider the real-world context in which models are deployed. Prospective validation, on the other hand, enables meaningful comparisons between data generation \emph{processes} by incorporating trained models and considering the subjective decisions that affect reproducibility. Prospective experiments are essential for consistent progress in modeling.
\end{abstract}


\end{frontmatter}

\section*{``You keep using that word''}

As the Man in Black clings to the Cliffs of Insanity in Rob Reiner's 1987 film \emph{The Princess Bride}, Vizzini exclaims, ``Inconceivable!'' Inigo Montoya responds: ``You keep using that word. I do not think it means what you think it means.'' If Inigo had studied chemistry rather than swordplay he might have said the same thing about `validation' of machine learning models and marveled at the disconnects that exist between model building and application.

Modeling in drug discovery is the pursuit of a moving target: the molecules we make (or buy) today often look different than the ones we made yesterday or that we'll make tomorrow. This introduces covariate shift~\cite{McGaughey2016-ob} into our datasets and requires continuous model updates to keep the applicability domain aligned with synthetic efforts. Temporal considerations have also become prominent in model evaluation, particularly with the introduction of time-split cross-validation~\cite{Sheridan2013-wu}.

A misunderstanding persists in the literature that temporal data splits are equivalent to prospective validation. In time-split cross-validation, models are trained on data generated before a certain date and tested on a held-out set of data generated after that date. However, this procedure makes an assumption about the data generation process that is violated by real-world application of the trained model: it does not consider the role of the model in compound selection.

In most programs, a complicated process integrating signals from medicinal chemists and predictive models is used to select the next batch of compounds for synthesis (more on that below). Introducing a new predictive model into this process will change the set of compounds that are synthesized such that the held-out set used to assess model performance is less representative of actual future data. Said another way, time-split cross-validation---like all other retrospective methods---can only tell you about how the model will perform \emph{if it is never used in the decision-making process}.

\section*{A Prospective Perspective}

True prospective validation uses the trained model to select compounds for testing. The model must have ``skin in the game'' to measure its effect on the data generation process~\cite{Bajorath2020-mv}. Depending on the process used to select compounds, prospectively incorporating the trained model may have different effects. For instance, the model may prioritize compounds that would not have been made in the original process or downvote molecules that would otherwise have been tested. The model could affect confidence estimates---perhaps the compound rankings are unchanged but predictions are more or less accurate than before.

It is critical to emphasize that prospective studies validate \emph{processes} and not models. While a predictive model may be part of a data generation process, it is not the only part: there are often choices to make regarding training data selection and preparation, input featurization, hyperparameter domains and optimization strategies, filtering or clustering of model output, integration of signals from chemists or other models, synthetic feasibility estimates, material costs and availability, and so on. All of these parameters affect the output of the process (i.e., the compounds that are selected for synthesis) and they may dilute or mask the perceived value of model-specific choices if they are not carefully controlled.

Many real-world processes include highly subjective components that make meaningful comparisons difficult. Consider a process that includes prioritization of model predictions by one or more medicinal chemists. If multiple chemists are involved, the effect of the change may not be consistent over repeated runs of the process; different chemists are likely to select different compounds~\cite{Lajiness2004-rz}, and the comparison could be repeated multiple times to get a distribution of expected outcomes. If one chemist does all of the selections, prospective validation should include selections from both the original and updated model predictions since individual preferences can change over time---the key principle is that process changes should be evaluated by comparisons starting \emph{from the same initial state}. Note that the validation results will be most representative of future performance if the same chemist(s) are also involved in downstream applications of the model.

\section*{From Cogs to Machines}

The impact of modeling changes may be attenuated or amplified by subjective decisions in the overall data generation process. Just as a model cannot perform better than the experimental error of the data, subjective process decisions limit the confidence we can have when assigning credit to modeling choices. In the worst cases, subjectivity can reduce otherwise meaningful comparisons to anecdotes.

One way to reduce subjectivity in prospective validation is to design systematic workflows that emphasize automation over manual engagement. For instance, prioritization of model predictions by a chemist could be replaced by predetermined filtering criteria or auxiliary models that remove undesirable compounds. McCloskey and colleagues recently took a step in this direction by employing ``automated or automatable filters''\footnote{The ``automatable'' filters were described thus: ``nonsystematic visual filtering was performed by a chemist with or without the aid of substructure searches that was restricted to removal of molecules with the potential for instability or reactivity''.} when selecting compounds to test models trained on DNA-encoded library selection data~\cite{McCloskey2020-es}. Automated decisions are also important components of closed-loop optimization systems~\cite{Parry2019-xl}.

The downside of automation is that it ignores ``chemical intuition'' and other decision-making criteria that are difficult to codify~\cite{Griffen2020-mf}. Fortunately, even if an automated workflow is not actually used in downstream applications, it can serve as a robust platform for prospective validation to the degree that automated decisions are reasonable proxies for their more subjective counterparts. This immediately suggests another experiment: prospective validation can be used to compare a real-world process to its automated analog! And of course, prospective validation can be used to evaluate changes to \emph{any} process regardless of whether it involves modeling.

\section*{Moving Forward}

We acknowledge that the financial and opportunity costs required for prospective experiments are not trivial and require larger-than-typical investments from organizations and teams. It would of course be impractical to demand that every model validation be prospective. Retrospective benchmarks like MoleculeNet~\cite{Wu2018-yc} are useful for broadly comparing new model architectures across many possible application domains. Competitions like the Merck Molecular Activity Challenge$^i$ create ``common task frameworks''~\cite{Donoho2017-fv} that reduce overfitting to test data. Task-specific retrospective validation (including time-split cross-validation) is helpful for identifying hyperparameter sensitivities and as a sanity check before investing additional resources into prospective studies.

Caveats aside, prospective validation is an essential tool for modern drug discovery. It provides the most realistic view of future performance by matching the actual setting for downstream applications and allows for direct comparisons with existing processes. As models are given more resources and a correspondingly larger role in discovery, systematic experiments that quantify and control for subjective variability enable accurate credit assignment to specific modeling decisions and other perturbations of data generation processes. It is this accurate credit assignment---enabled by true prospective validation---that will set the stage for a future where chemists and modelers work together to create \emph{reproducible discovery processes} that provide clear signals for optimization.

Inigo Montoya's all-consuming quest to avenge the murder of his father by the six-fingered man eventually came to an end, but our quest is just beginning. As a field, we must adopt a prospective perspective. This includes (i) holding up prospective validation as the most reliable way to estimate model performance; (ii) acknowledging the incompleteness of retrospective validation in our own work and peer reviews; and (iii) adjusting our data generation processes and validation workflows to give robust, reproducible results. Only in this way can we avenge ourselves of anecdotal reports and apples-to-oranges comparisons that hold us back from consistent progress.

\section*{Glossary}

\textbf{Time-split cross-validation}: A method described by Sheridan~\cite{Sheridan2013-wu} for estimating model performance that uses a single point in time to split the available data into training and test sets.

\textbf{Prospective validation}: A method for estimating model performance that simulates the anticipated real-world application context, including the effects that the model may have on the acquisition and statistical distribution of future data.

\textbf{Data generation process}: A sequence of steps that results in the acquisition of new data. A typical data generation process may include the prediction of molecular properties as well as choices about input distribution and representation, post-prediction filters, etc.

\textbf{Hyperparameter domain}: The type and range of meta-parameters (e.g. the number of trees in a random forest) considered when optimizing a predictive model. The set of hyperparameter values explored during optimization (usually against a held-out tuning data set) may limit the performance of the final model.

\section*{Acknowledgments}

S.K. is grateful to Patrick Riley, Alex Wiltschko, and Brian Lee for helpful discussions and feedback.

\section*{Resources}

\begin{enumerate}[label=i.,leftmargin=*]
    \small
    \item Kaggle, Merck molecular activity challenge (2012). URL https://www.kaggle.com/c/MerckActivity.
\end{enumerate}

\bibliography{trends}

\end{document}